\documentclass{article}

    \PassOptionsToPackage{numbers, compress}{natbib}

\usepackage[preprint]{neurips_2024}

\usepackage[utf8]{inputenc} 
\usepackage[T1]{fontenc}    
\usepackage{hyperref}       
\hypersetup{
    colorlinks = true,
    citecolor = {teal},
}
\usepackage{url}            
\usepackage{booktabs}       
\usepackage{amsfonts}       
\usepackage{nicefrac}       
\usepackage{microtype}      
\usepackage[dvipsnames]{xcolor}         

\usepackage{amsmath}
\usepackage{amsfonts}
\usepackage{graphicx}
\usepackage{multirow}
\usepackage{caption}
\usepackage{wrapfig}

\usepackage{pifont}
\newcommand{\cmark}{\ding{51}}%
\newcommand{\xmark}{\ding{55}}%

\usepackage{colortbl}
\definecolor{mygray}{gray}{.95}

\newcommand{\oursf}{Adapter-\textit{F}}
\newcommand{\oursi}{Adapter-\textit{I}}

\newcommand{\oursff}{WMAdapter-\textit{F}}
\newcommand{\oursii}{WMAdapter-\textit{I}}

\title{WMAdapter: Adding WaterMark Control to Latent Diffusion Models}

\author{%
  Hai Ci \quad Yiren Song \quad Pei Yang \quad Jinheng Xie \quad Mike Zheng Shou \\
  Show Lab, National University of Singapore \\
  \texttt{\{cihai03, sierkinhane, mike.zheng.shou\}@gmail.com} \\
  \texttt{yiren@nus.edu.sg, yangpei@u.nus.edu} \\
}

\begin{document}

\maketitle

\begin{abstract}
  Watermarking is crucial for protecting the copyright of AI-generated images. We propose WMAdapter, a diffusion model watermark plugin that takes user-specified watermark information and allows for seamless watermark imprinting during the diffusion generation process. WMAdapter is efficient and robust, with a strong emphasis on high generation quality. To achieve this, we make two key designs: (1) We develop a contextual adapter structure that is lightweight and enables effective knowledge transfer from heavily pretrained post-hoc watermarking models. (2) We introduce an extra finetuning step and design a hybrid finetuning strategy to further improve image quality and eliminate tiny artifacts. 
  Empirical results demonstrate that WMAdapter offers strong flexibility, exceptional image generation quality and competitive watermark robustness.
\end{abstract}
\section{Introduction}

With the widespread adoption of diffusion models~\citep{ho2020denoising,podell2023sdxl,song2020score,Rombach_2022_CVPR,ci2023gfpose,zhang2023show}, diffusion-generated images are proliferating across media and the internet. While these models meet the demand for high-quality creative content, their misuse raises significant concerns about copyright protection and the security of images against deepfakes~\citep{westerlund2019emergence}. Watermarking technology~\citep{cox2007digital} provides a tailored solution for resolving copyright disputes and identifying the sources of forgeries.

Previous watermarking methods added watermarks to images in a post-hoc way through frequency domain transformations~\citep{cox2007digital,lin2001rotation,xia1998wavelet} or by training watermark encoding networks~\citep{zhu2018hidden,tancik2020stegastamp,zhang2019robust}. However, in the context of watermarking diffusion images, post-hoc methods introduce additional workflows and separate from diffusion pipeline. Recently, more efforts~\citep{zhao2023recipe,fernandez2023stable,min2024watermark,xiong2023flexible,lei2024diffusetrace,meng2024latent,yang2024gaussian,ci2024ringid} have focused on leveraging the characteristics of the diffusion process to seamlessly integrate watermarking into the diffusion pipeline, known as diffusion-native watermarking.
Among these, Stable Signature~\citep{fernandez2023stable} proposed a method that fine-tunes the VAE decoder of a latent diffusion model~\citep{Rombach_2022_CVPR} using a pretrained watermark decoder~\citep{zhu2018hidden}. 
This approach has shown promising results. However, it requires fine-tuning a different VAE decoder for each unique watermark, making it difficult to scale to millions of keys necessary for real large-scale scenarios. Additionally, the tuning of VAE decoder on a small amount of data results in blurrier generated images and lens flare-like artifacts (see Fig.~\ref{fig:cmp_stablesig}).

To address these issues, we propose WMAdapter, a diffusion watermark plugin (Fig.~\ref{fig:framework}) that accepts arbitrary user input watermark bits and imprints the watermark on-the-fly without per-watermark finetuning. 
Since WMAdapter serves as a plug-and-play module and keeps the original VAE decoder intact, it can also produce sharper images with minimal artifacts.

Many concurrent works~\citep{bui2023rosteals,xiong2023flexible,min2024watermark,meng2024latent,zhang2024training,kim2023wouaf} also explores watermark plugins for diffusion models. However, their approaches typically involve lengthy training pipeline or heavy structures, such as finetuning a heavy UNet~\citep{ronneberger2015u} plugin at each denoising time step~\citep{min2024watermark}, training the plugin and watermark decoder from scratch~\cite{kim2023wouaf,xiong2023flexible}, or using a heavy watermark decoder~\citep{bui2023rosteals}. In contrast, our motivation is to create a solution that is easy to train and use for diffusion practices. 
To achieve this, we have designed a lightweight structure, utilizing a pretrained watermark decoder and a minimal training pipeline. Tab.~\ref{tab:introduction_cmp} compares several concurrent methods. 
This lightweight design also poses significant challenges in achieving a robust and high-quality watermark adapter. To be both efficient and powerful, we propose a novel \textbf{Contextual Adapter} structure for efficient and effective knowledge transfer from pretrained decoders. 
Unlike widely adopted non-contextual adapters~\citep{bui2023rosteals,xiong2023flexible,kim2023wouaf,ye2023ip} that only condition on the watermark bits, our contextual structure also conditions on the cover image features from the VAE (hence "contextual"). 
We found this dual conditioning better adapts watermark to the generated images, being crucial for high-quality knowledge transfer and image generation.

\vspace{-0.2cm}
\begin{table}[ht]
  \caption{Comparison between diffusion watermark plugins. \textcolor{orange}{"Orange"} colors modified modules. ``WM Dec'' is short for ``Watermark Decoder''. WADIFF~\citep{min2024watermark} and AquaLoRA~\citep{feng2024aqualora} alter the layout of generated images. FSW~\citep{xiong2023flexible} has to work with a finetuned VAE decoder.}
  \label{tab:introduction_cmp}
  \centering
  \begin{tabular}{lllcc}
    \toprule
     & Plug onto & Modules Involved in Training & Plug-and-play & Imperceptible  \\
    \midrule
    WADIFF~\citep{min2024watermark} & UNet &\textcolor{orange}{Plugin}+UNet+VAE+WM Dec &\textcolor{OliveGreen}{\cmark} &\textcolor{OrangeRed}{\xmark} \\
    AquaLoRA~\citep{feng2024aqualora} & UNet &\textcolor{orange}{Plugin}+\textcolor{orange}{UNet}+VAE+WM Dec &\textcolor{OliveGreen}{\cmark} &\textcolor{OrangeRed}{\xmark} \\
    FSW~\citep{xiong2023flexible} & VAE &\textcolor{orange}{Plugin+VAE+WM Dec}  &\textcolor{OrangeRed}{\xmark} &\textcolor{OliveGreen}{\cmark} \\
    WOUAF~\citep{kim2023wouaf} & VAE &\textcolor{orange}{Plugin+VAE+WM Dec}  &\textcolor{OliveGreen}{\cmark} &\textcolor{OliveGreen}{\cmark} \\
    \rowcolor{mygray}
    Ours & VAE &\textcolor{orange}{Plugin}+VAE+WM Dec  &\textcolor{OliveGreen}{\cmark} &\textcolor{OliveGreen}{\cmark} \\
    \bottomrule
  \end{tabular}
\end{table}

To further improve the image generation quality, we introduce an extra finetuning stage and propose a novel \textbf{Hybrid Finetuning} strategy. This strategy involves jointly finetuning the Adapter and the VAE decoder, while using the original VAE decoder during inference. It effectively suppresses tiny artifacts and significantly enhances image sharpness.
We summarize  our contributions as follows: 
\begin{enumerate}
    \item We introduce WMAdapter, a novel framework that dynamically  imprints different watermarks for diffusion models without requiring individual finetuning for each watermark. 
    \item We propose a compact contextual adapter structure and a hybrid finetuning strategy, ensuring efficient learning and high quality watermarking. 
    \item Experimental results show that WMAdapter not only maintains watermark robustness but also significantly enhances flexibility, scalability, and image generation quality, 
\end{enumerate}

\section{Related work}

\subsection{Post-hoc watermarking}
Post-hoc methods include traditional frequency domain transformation methods such as DwtDct~\citep{cox2007digital} and DwtDctSvd~\citep{cox2007digital}, per-image optimization-based methods like SSL~\citep{fernandez2022watermarking} and FNNS~\citep{kishore2021fixed}, and forward encoding/decoding methods such as HiDDeN~\citep{zhu2018hidden}, Stegastamp~\citep{tancik2020stegastamp}, MBRS~\citep{jia2021mbrs}. Different methods have various design focuses. For instance, FNNS emphasizes hiding more bits, HiDDeN and MBRS prioritizes robustness against JPEG compression, and Stegastamp targets robustness against real-world re-photography.

\subsection{Diffusion-native watermarking}
According to the location of the watermark, we classify diffusion-native watermarking methods into two categories.
\textbf{Adding to initial noise:}
Tree-Ring~\citep{wen2023tree} proposed adding watermarks to the diffusion initial noise, achieving remarkable robustness. However, this approach lacks multi-key identification capabilities~\citep{ci2024ringid}. Subsequent methods improve by using enhanced imprinting techniques~\citep{yang2024gaussian,ci2024ringid}, or employing encoder/decoder on initial noise~\citep{lei2024diffusetrace}. However, these methods significantly alter the layout of the generated images, which is not desirable in some production scenarios.
\textbf{Adding to latent space:}
This category involves imprinting watermarks into the latent (or feature) space of the VAE or the diffusion UNet, with most methods not altering the generated image layout. Methods like RoSteALS~\citep{bui2023rosteals}, LW~\citep{meng2024latent}, and PAP~\citep{zhang2024training} imprint watermarks into the VAE's latent space, requiring more effort to train both the watermark plugin and the watermark decoder from scratch. RoSteALS and LW also suffer from unstable training, necessitating multi-stage training or subtle hyperparameter tuning.
WADIFF~\citep{min2024watermark} and AquaLoRA~\citep{feng2024aqualora} imprints watermarks into the UNet backbone, resulting in longer training pipelines and changes to the generated image layout. FSW~\citep{xiong2023flexible}, StableSignature~\citep{fernandez2023stable}, and WOUAF~\citep{kim2023wouaf} are similar to our approach in that they inject watermarks into the VAE feature space, but they need to modify the VAE decoder, which tends to degrade the generation quality. 
In contrast, WMAdapter keeps the entire diffusion pipeline intact and can be directly plugged onto the VAE decoder (Fig.~\ref{fig:framework}), providing greater flexibility and higher generation quality.

\section{Method}
\label{sec:method}
In this section, we will introduce the framework of WMAdapter, detail its contextual structure, and discuss the training and fine-tuning strategies.

\subsection{Framework overview}
\begin{figure}[t]
  \centering
  \includegraphics[width=\linewidth]{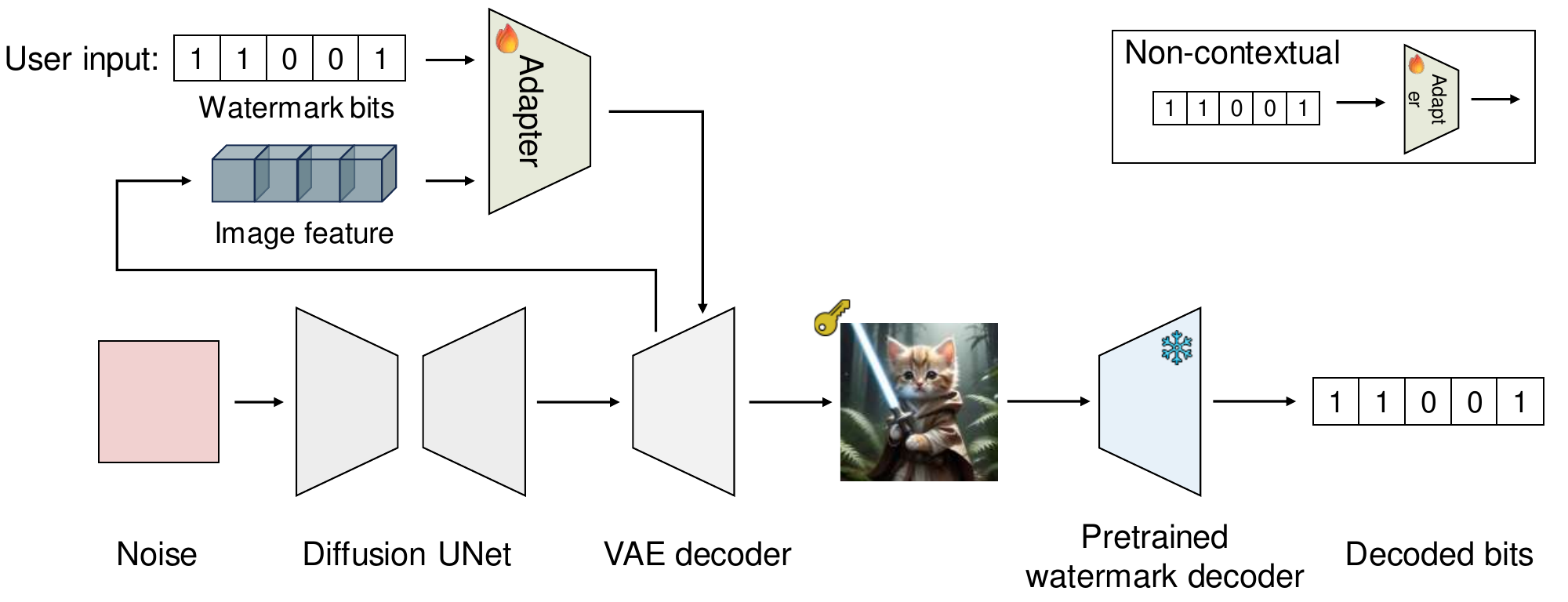}
   \caption{Framework overview. WMAdapter is plugged onto the VAE decoder. It takes user input watermark bits and image features from the VAE decoder, imprinting the watermark on-the-fly during VAE decoding. In contrast, traditional non-contextual adapters  take only watermark conditions as input. WMAdapter can be trained with a post-hoc watermark decoder for efficient knowledge transfer.  The image and icons credit to~\citep{sdonline,flaticon}.}
   \vspace{-0.2cm}
   \label{fig:framework}
\end{figure}

Fig.\ref{fig:framework} illustrates the overall framework of WMAdapter. WMAdapter is a plug-and-play watermark module that can be directly attached to the VAE decoder of a latent diffusion model~\citep{Rombach_2022_CVPR}. It imprints the watermark during image generation, seamlessly integrating into the diffusion generation workflow. WMAdapter employs a novel contextual adapter structure, which takes both watermark bits and image features from the VAE decoder as input and outputs feature residuals containing watermark information. 
Watermarked images can be directly fed into a pretrained watermark decoder, such as HiDDeN~\citep{zhu2018hidden}, to retrieve the watermark information.


The training of WMAdapter consists of two stages: large-scale training and fast finetuning. In the training stage, we freeze the VAE decoder and the watermark decoder and train only the Adapter on a large scale dataset. 
We then finetune the Adapter and VAE decoder on a small amount of data.
Specifically, we present a novel hybrid finetuning strategy that is able to suppress tiny artifacts and significantly enhance generation quality. We also discuss several different strategies concerning different tradeoffs between robustness and quality. 


\subsection{Contextual adapters}
\begin{figure}[t]
  \centering
  \includegraphics[width=\linewidth]{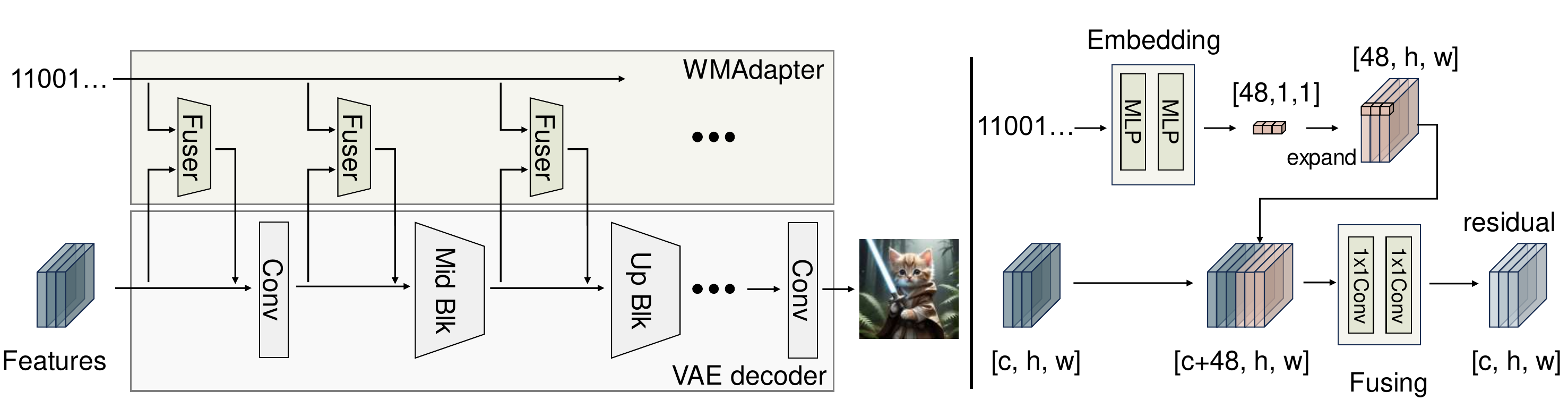}
   \caption{The architecture of WMAdapter. \textit{Left:} The structure of WMAdapter. It comprises several independent Fusers with identical structures. \textit{Right:} The structure of Fuser. It consists of a watermark Embedding module and a Fusing module.}
   \vspace{-0.2cm}
   \label{fig:contextual_adapter}
\end{figure}

In this section, we provide a detailed overview of the contextual structure of WMAdapter. Fig.~\ref{fig:contextual_adapter} (\textit{Left}) illustrates the internal structure of WMAdapter, which comprises a series of independent {\it Fuser} modules. Each {\it Fuser} $\phi_i(\cdot)$ is attached before a corresponding VAE decoder block $i$. It receives both VAE feature $f_i$ and watermark bits $w$ as inputs, and outputs a feature residual $y_i$ to update $f_i$. Formally,
\begin{equation}
    \begin{aligned}
    y_i &= \phi_i\left(f_i, w\right), \\
    f_i' &= f_i + y_i.
    \end{aligned}
\end{equation}
We put a total of 6 {\it Fusers} before the Conv Block, Middle Block and four Up Blocks in the kl-f8 VAE decoder used by Stable Diffusion~\cite{Rombach_2022_CVPR}. 

Fig.~\ref{fig:contextual_adapter} (\textit{Right}) illustrates the internal structure of an Fuser. An Fuser consists of two main components: the Embedding module and the Fusing module. The Embedding module maps the 01 bit sequence into a 48-dimensional watermark feature vector. This feature vector is then expanded along the width and height dimensions to produce a watermark feature map with the same dimensions as the image feature. The image feature and watermark feature are concatenated along the channel dimension and fed into the Fusing module, which outputs the image feature residuals. 
Keeping lightweight in mind, we use two MLPs with 256 intermediate feature channels for the Embedding module, and two 1x1 convolutions with half the image feature channels $\frac{c}{2}$ as intermediates for the Fusing module. We employ LeakyReLU as the non-linearity. The total parameters of WMAdapter are only 1.3M, making it a small and efficient plugin.

\subsection{Training}
\label{subsec:training}
In the training stage, we use a pretrained watermark decoder to decode watermark bits from the watermarked images. We freeze the watermark decoder and the VAE decoder, and only train the Adapter. Why do we use a pretrained decoder instead of training a watermark decoder from scratch along with the Adapter? We observe that training an encoder/decoder pair from scratch, as post-hoc methods do, typically requires significant training effort. For example, HiDDeN takes 300 epochs to converge on the COCO dataset. Using a pretrained post-hoc decoder facilitates efficient knowledge transfer, allowing WMAdapter to converge in just 1-2 epochs.
We use two types of losses as our objective: the consistency loss between the watermarked image $x_w$ and the unwatermarked image $x$, and the accuracy of decoded bits. The total loss function is defined as:

\begin{equation}
    \mathcal{L} = \lambda_1 \mathcal{L}_{mae}\left(x, x_w\right) + \lambda_2 \mathcal{L}_{lpips}\left(x, x_w\right) + \lambda_3 \mathcal{L}_{vgg}\left(x, x_w\right) +  \lambda_4 \mathcal{L}_{bce}\left(w, w'\right)
\end{equation}

where the first three terms represent image consistency losses. We use MAE and LPIPS loss~\citep{zhang2018unreasonable} to maintain consistency with VAE pretraining~\citep{Rombach_2022_CVPR}. Additionally, we include a Watson-VGG loss~\citep{czolbe2020loss} similar to Stable Signature~\citep{fernandez2023stable} to enhance human visual preference. For watermark decoding accuracy, we use binary cross-entropy loss bewteen decoded bits $w'$ and input bits $w$. We empirically set $\lambda_1, \lambda_2, \lambda_3, \lambda_4$ to 0.2, 0.2, 0.08, 1.0, respectively.


\subsection{Hybrid finetuning}
\label{subsec:finetuning}

After the training stage, we obtain a watermark adapter that performs well in both accuracy and image quality (Sec.~\ref{subsubsec:ablation_finetune}). However, when we zoom in on the generated images by 20 times, grid-like artifacts can often be observed (Fig.~\ref{fig:ablation_finetuning}). To further improve image quality and eliminate these tiny artifacts, we introduce a fine-tuning stage with a small amount of data. In addition to the first stage training losses, we incorporate a total variation loss~\citep{tvloss} on the watermarked images to enhance smoothness, setting its weight to 0.02.

\begin{figure}[t]
  \centering
  \includegraphics[width=\linewidth]{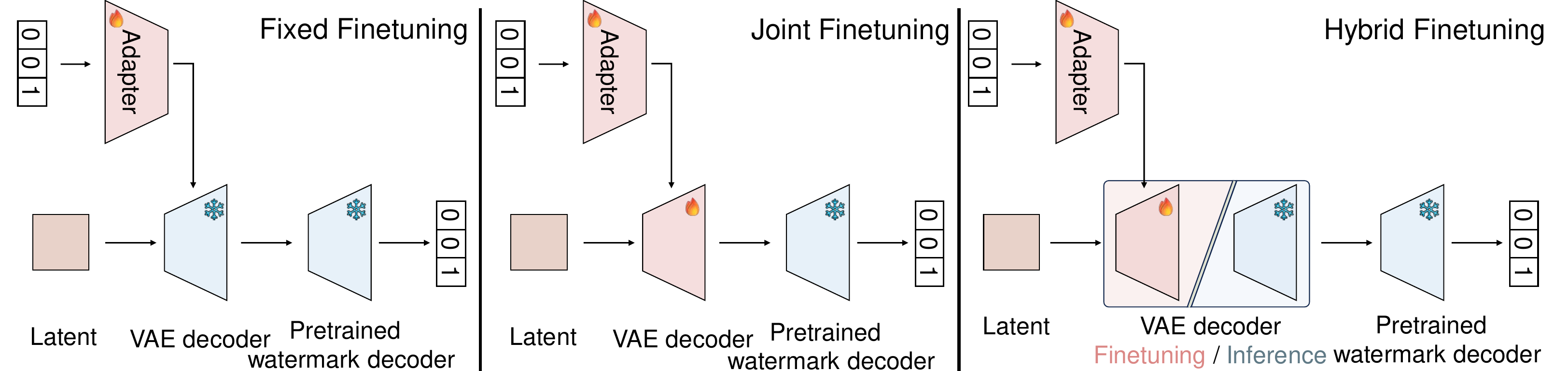}
   \caption{Illustration of 3 different finetunig strategies. They differ in how to treat the VAE decoder.}
   \vspace{-0.2cm}
   \label{fig:hybrid_finetuning}
\end{figure}

Additionally, we propose a novel Hybrid Finetuning strategy. We finetune both the Adapter and the VAE decoder, but use the fine-tuned Adapter and the original VAE decoder for inference. Fig.~\ref{fig:hybrid_finetuning} distinguishes this method from two other classic finetuning strategies: Fixed and Joint Finetuning.
The Fixed Finetuning strategy uses the same training approach as in the first stage, fixing the VAE decoder and quickly finetuning the Adapter with a high learning rate. The Joint Finetuning strategy jointly finetunes the Adapter and the VAE decoder, using both finetuned copies for inference. 

Empirical results in Sec.~\ref{subsubsec:ablation_finetune} demonstrate that Hybrid Finetuning effectively removes tiny artifacts and, by keeping the VAE intact, produces the sharpest and clearest images while maintaining the plug-and-play advantage, making it ideal for scenarios requiring high image quality. The Fixed Finetuning strategy improves image quality compared to the first stage, but it fails to effectively eliminate tiny artifacts. Joint Finetuning can slightly enhance robustness, but the modifications to the VAE decoder result in blurrier and smoother images, as well as lens flare-like artifacts.


\subsection{Discussion}
The design of WMAdapter emphasizes lightweight, efficient, and flexible operation. We prioritize watermark invisibility while maintaining watermark robustness. This contrasts with previous methods that focused more on robustness at the expense of image quality or simplicity. 
We argue that high image quality, along with light weight, flexibility, and ease of use should be the top priorities when deploying watermarking technology in diffusion products.
One evidence may be that the Stable Diffusion series~\citep{Rombach_2022_CVPR,podell2023sdxl} use traditional DwtDct~\citep{cox2007digital} as the default watermarking method. This focus is the rationale behind WMAdapter's lightweight design and hybrid finetuning strategy.
\section{Experiment}
\label{sec:experiment}

\subsection{Experimental setup}
\paragraph{Model and dataset}
We experiment with a popular latent diffusion model Stable Diffusion 2.1~\citep{Rombach_2022_CVPR} and its associated kl-f8 VAE. We adopt the pretrained watermark decoder from HiDDeN~\citep{zhu2018hidden}. The checkpoint we use was pretrained by~\citep{fernandez2023stable}, encoding 48-bits watermark information. This checkpoint is also used to finetune Stable Signature~\citep{fernandez2023stable}. Thus, our adapter can be  directly compared with~\citep{fernandez2023stable}. ALL training and finetuning steps are performed on MS-COCO 2017~\citep{lin2014microsoft} training set. Validation is performed on COCO 2017 validation set. We train and evaluate our adapters on images at resolution $512\times512$. For images smaller than this size, we resize their shorter edge to 512, then center crop to get a $512\times512$ image. 

\paragraph{Training strategies}
For the first stage training, we adopt 8 $\times$ NVIDIA A5000 GPUs of 24 GB memory, with per-GPU batchsize of 2, AdamW optimizer~\citep{loshchilov2017decoupled}, a learning rate of 5e-4. We train the model for 2 epochs, taking about 5 hours. In comparison, post-hoc methods like HeDDeN~\citep{zhu2018hidden} usually take hundreds of epochs to converge, showing the value of adapting from pretrained models.

For the second stage finetuning, we use a single A5000 GPU. We set the mini-batch to 2. We also use the AdamW optimizer and a start learning rate of 5e-4. However, we adopt a per-step cosine learning rate decay with 20 warm-up steps. Unless otherwise specified, the total 
fine-tuning process defaults to 2,000 steps, lasting for about 50 minutes.
Different finetuning strategies result in several different adapter variants. We use Adapter-{\it B}, Adapter-{\it F}, Adapter-{\it V}, and Adapter-{\it I} to denote the adapters obtained by No Finetuning, Fixed Finetuning, Joint Finetuning and Hybrid Finetuning, respectively. 

\paragraph{Evaluation metric}
Following previous conventions~\citep{zhu2018hidden,fernandez2022watermarking,fernandez2023stable}, we use average bit accuracy to evaluate the watermarking performance of our adapter. Bit accuracy is defined by the ratio of correctly decoded bits in a 48-bit watermark sequence. 
Apart from the bit accuracy, we also report the tracing accuracy among different numbers of users following concurrent works~\citep{min2024watermark,ci2024ringid,yang2024gaussian}. We adopt the evaluation protocol of~\citep{min2024watermark}. 
Concretely, we construct user pools of different sizes, ranging from $10^4$ to $10^6$, to evaluate the accuracy of user tracing at different scales. Each user is assigned a unique key.
For each user pool. we randomly select 1,000 users and watermark 5 images per user, resulting in 5,000 watermarked images. For each of the 5,000 images, we find the best match among the user pool and check if it's a correct match. Tracing accuracy is then averaged over all 5,000 images. 

In addition to accuracy measurements, we are also interested in the watermark’s invisibility and image generation quality. We report the Peak Signal-Noise-Ratio (PSNR), Structural Similarity Index Measure (SSIM)~\citep{wang2004image}, and Fréchet Inception Distance (FID)~\citep{heusel2017gans} metrics between images before and after watermarking. Typically, higher PSNR and SSIM means more similar to the original image, with less noise and distortion, leading to sharper and clearer images. While lower FID means the watermarked images have higher fidelity and more closely resemble the real images in terms of appearance and variety.

\begin{table}
  \caption{Comparison with other watermarking methods on generation quality and robustness. All methods are evaluated on COCO 2017 val set~\citep{lin2014microsoft} with image size $512 \times 512$. Since Stable Signature~\citep{fernandez2023stable} requires finetuning of separate VAE decoders to embed different keys, we report its average results on 10 randomly sampled keys. For robustness, we use Crop 0.3, JPEG 80, Brightness 1.5. Following~\citep{fernandez2023stable}, we use (Crop 0.5, Jpeg 80, Brightness 1.5) for combined attacks.}
  \label{tab:cmp_bit_acc}
  \centering
  \begin{tabular}{llllllllll}
    \toprule
    & & & & & \multicolumn{5}{c}{Bit Accuracy $\uparrow$} \\
    \cmidrule(r){6-10}
    &Method &PSNR $\uparrow$ &SSIM $\uparrow$ &FID $\downarrow$ &None &JPEG &Crop &Bright &Comb \\
    \midrule
    \multirow{4}{*}{\rotatebox[origin=c]{90}{\textit{Post-hoc}}} & DwtDct~\citep{cox2007digital} & \textbf{39.1} & \underline{0.97} & \underline{2.6} & 0.79 & 0.50 & 0.51 & 0.79  & 0.50 \\
    & DwtDctSvd~\citep{cox2007digital} & \textbf{39.1} & \textbf{0.98} & 3.2 & 0.72 & 0.71 & 0.50 & 0.49  & 0.50 \\
    & SSL~\citep{fernandez2022watermarking} & 33.0 & 0.89 & 14.8 & \textbf{1.00} & \textbf{0.99} & 0.97 & 0.98  & 0.92 \\
    & HiDDeN~\citep{zhu2018hidden} & 34.1 & 0.95 & 3.1 & 0.98 & 0.84 & 0.97 & 0.98  & 0.87 \\
    \midrule
    \multirow{3}{*}{\rotatebox[origin=c]{90}{\textit{Merged}}} & Stable Signature~\citep{fernandez2023stable} & 29.7 & 0.87 & 3.2 & \underline{0.99} & \underline{0.93} & \textbf{0.99} & \textbf{0.99}  & \textbf{0.94} \\
    & \cellcolor{mygray}WMAdapter-F & \cellcolor{mygray}33.1 & \cellcolor{mygray}0.95 & \cellcolor{mygray}2.7 & \cellcolor{mygray}\underline{0.99} & \cellcolor{mygray}0.92 & \cellcolor{mygray}\textbf{0.99} & \cellcolor{mygray}\textbf{0.99}  & \cellcolor{mygray}\underline{0.93} \\
    & \cellcolor{mygray}WMAdapter-I & \cellcolor{mygray}\underline{34.8} & \cellcolor{mygray}0.96 & \cellcolor{mygray}\textbf{2.5} & \cellcolor{mygray}0.98 & \cellcolor{mygray}0.90 & \cellcolor{mygray}0.97 & \cellcolor{mygray}0.97  & \cellcolor{mygray}0.91 \\
    \bottomrule
  \end{tabular}
\end{table}

\subsection{Comparison with other methods}
\paragraph{Bit accuracy and image quality}
We compare our method with four post-hoc watermarking methods~\citep{cox2007digital,fernandez2022watermarking,zhu2018hidden} and one merged diffusion watermarking method Stable Signature~\citep{fernandez2023stable}. Among the post-doc methods, DwtDct~\citep{cox2007digital} and DwtDctSvd~\citep{cox2007digital} are traditional frequency-based watermarking methods. DwtDct is also the default watermarking method used by Stable Diffusion~\citep{Rombach_2022_CVPR}. SSL~\citep{fernandez2022watermarking} and HiDDeN~\citep{zhu2018hidden} are neural network-based methods. SSL is an iterative optimization method, while HiDDeN is an encoder-decoder method. For HiDDeN, we use the model provided by~\citep{fernandez2023stable}.

Tab.~\ref{tab:cmp_bit_acc} shows the comparison. 
Traditional methods DwtDct and DwtDctSvd achieve the highest PSNR and SSIM, at 39.1 dB and 0.98 respectively, indicating their ability to generate high-quality images with high watermark invisibility. However, these methods exhibit relatively weak robustness and cannot handle common image transformations such as cropping and JPEG compression. Even on clean images, they achieve only about 0.7-0.8 bit accuracy.

Neural network-based methods like SSL, HiDDeN, and Stable Signature are significantly more robust against various attacks due to training time augmentation. They achieve $0.9$ or higher bit accuracy under JPEG compression, cropping, brightness adjustment, and combined attacks. However, this robustness comes at the cost of image quality. SSL produces noticeable artifacts on generated images (Fig.~\ref{fig:appendix_difference_others}) and results in a very high FID. Stable Signature, on the other hand, produces smoother but blurrier images due to the fine-tuning of the VAE decoder, achieving a PSNR of 29.7 dB, an SSIM of 0.87, and an FID of 3.2. We also observe that Stable Signature tends to produce lens flare-like artifacts. This is shown in Fig.~\ref{fig:cmp_stablesig} and will be discussed in Sec.~\ref{subsec:qualitative}.

Our adapters achieve the best image quality among neural network-based methods, excelling in PSNR, SSIM, and FID metrics. \oursff~achieves a PSNR of 33.1 dB, an SSIM of 0.95, and an FID of 2.7, significantly outperforming Stable Signature while maintaining comparable robustness. Notably, Stable Signature requires fine-tuning the VAE decoder for each specific watermark key, making it impractical for scaling to millions of keys. In contrast, our adapter can embed up to $2^{48}$ different keys on the fly.
Employing a hybrid fine-tuning strategy, \oursii~further improves PSNR, SSIM, and FID to 34.8 dB, 0.96, and 2.5, at a little cost of 0.01-0.02 bit accuracy. Its FID score even surpasses the traditional methods DwtDct and DwtDctSvd. Qualitative comparisons in Sec.~\ref{subsec:qualitative} also show that \oursii~produces sharper images with minimal artifacts. Overall, \oursii~strikes good balance between robustness and generation quality, making it suitable for scenarios where high image quality is prioritized..

\vspace{-0.2cm}

\begin{figure}[t]
    \centering
    \begin{minipage}[H]{0.53\linewidth}
        \centering
        \captionof{table}{Accuracy of tracing different numbers of keys. All methods are evaluated on COCO dataset~\citep{lin2014microsoft}. For WADIFF$^{\ast}$~\citep{min2024watermark}, we use the number reported by its original paper.}
        \setlength{\tabcolsep}{1.4pt}
        \resizebox{\linewidth}{!}{%
        \begin{tabular}{llll}
            \toprule
            Method & Trace $10^4$ & Trace $10^5$ & Trace $10^6$ \\
            \midrule
            WADIFF$^{\ast}$~\citep{min2024watermark} & 0.982 & 0.968 & 0.934 \\
            Tree-Ring~\citep{wen2023tree} & 0.000 & 0.000 & 0.000 \\
            Stable Signature~\citep{fernandez2023stable} & 0.999 & 0.999 & 0.998 \\
            \rowcolor{mygray}
            \oursff & \textbf{1.000} & \textbf{1.000} & \textbf{1.000} \\
            \rowcolor{mygray}
            \oursii & \textbf{1.000} & 0.999 & 0.999 \\
            \bottomrule
      \end{tabular}}
      \label{tab:cmp_id_acc}
    \end{minipage}
    \hspace{0.02\textwidth}
    \begin{minipage}[H]{0.43\linewidth}
        \centering
        \includegraphics[width=\linewidth]{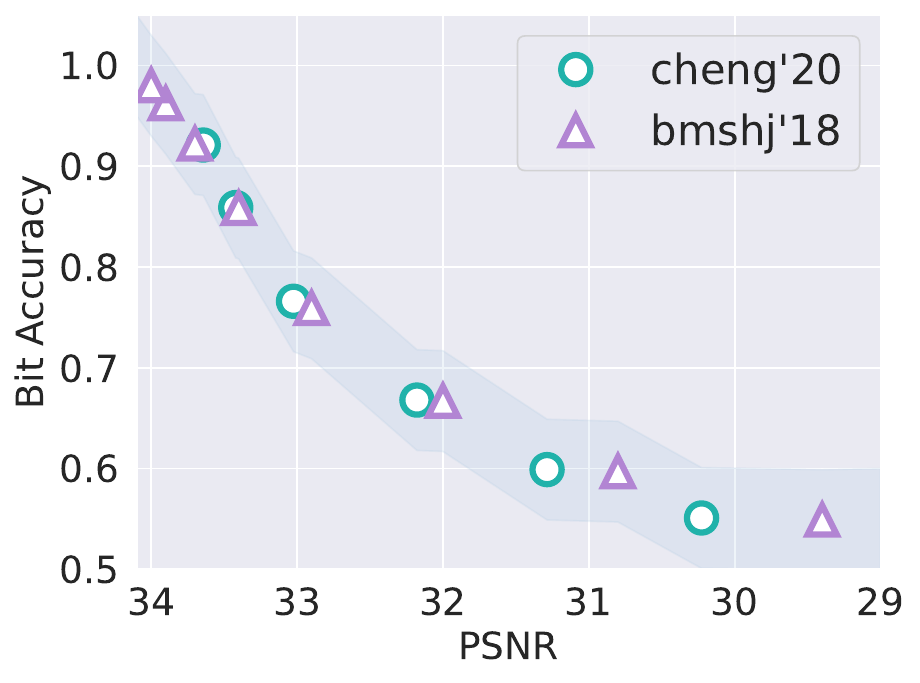}
        \caption{\oursff~against auto-encoder watermark removal~\citep{ballemshj18,cheng2020image}.}
        \label{fig:ablation_vae_attack}
    \end{minipage}
\end{figure}

\paragraph{Tracing accuracy}
We further compare the tracing accuracy with 3 diffusion-native watermarking methods~\citep{min2024watermark,wen2023tree,fernandez2023stable} in Tab.~\ref{tab:cmp_id_acc}. We can see that our adapters achieve nearly perfect tracing accuracy with different scales of users.
Tree-Ring~\citep{wen2023tree} achieves zero tracing accuracy due to its design flaws uncovered by~\citep{ci2024ringid}. WADIFF~\citep{min2024watermark} is a concurrent effort, which also employs HiDDeN decoder to finetune
a watermark plugin for diffusion models. Since it also evaluates on COCO dataset, we report its numbers as a reference.
We can see that its tracing accuracy gradually drops as the scale grows despite they employ a heavier adapter and longer training pipeline. Both ours and Stable Signature perform consistently at different user scales. Notably, Stable Signature has higher average bit accuracy but gets slightly worse tracing accuracy than our adapters. We attribute this to its larger performance variance among different keys.

\vspace{-0.1cm}

\subsection{Robustness to more attacks}
\vspace{-0.1cm}

\paragraph{Various attack intensities}
Fig.~\ref{fig:appendix_range} shows the robustness of our adapters against a range of attack intensities. Our adapters achieve comparable performance to Stable Signature under various levels of attacks, while offering flexibility, scalability and higher image quality.

\vspace{-0.1cm}

\paragraph{Against auto-encoders}
We evaluate our \oursff against compression attacks by two neural auto-encoders~\citep{cheng2020image,ballemshj18} at different compression rates. We use models provided by~\citep{begaint2020compressai}. 
For cheng'20~\citep{cheng2020image} and bmshj'18~\citep{ballemshj18}, we evaluate at compression rate 1-6 and 1-8, respectively.
Fig.~\ref{fig:ablation_vae_attack} shows the Accuracy-PSNR curve. To successfully reduce bit accuracy to around 0.5, a strong compression rate has to be used, resulting in approximately a 5 dB loss in PSNR, showing similar trend to~\citep{fernandez2023stable,kim2023wouaf}.

\subsection{Ablation Study}

\subsubsection{Why contextual structure?}
We study the effects of different adapter structures. Tab. \ref{tab:ablation_adapter_structure} shows the comparison between different adapter variants after the first stage training. If we use $3\times3$ conv kernels instead of $1\times1$ kernels to fuse image features and watermark bits, the adapter suffers from unstable training.
If the adapter is non-contextual, {\it i.e.}, only takes user bit sequence as input and does not receive image features, the bit accuracy on clean images will decrease by 0.02, and the PSNR will significantly drop by 4.1 dB to 28.7 dB. This demonstrates the importance of image features for watermark accuracy and invisibility. This is a crucial observation. We have found that many concurrent watermarking efforts still use the non-contextual plugin architecture~\citep{min2024watermark,xiong2023flexible,kim2023wouaf,bui2023rosteals}. Our observation provides a simple yet promising approach to further enhance these methods.


\begin{wraptable}{R}{7.5cm}
  \vspace{-0.3cm}
  \caption{Ablation study on adapter structures.}
  \label{tab:ablation_adapter_structure}
  \centering
  \begin{tabular}{llll}
    \toprule
     & Ours & Non-contextual & Conv $3\times3$ \\
    \midrule
    Bit Acc &\textbf{0.99} &0.97 &0.49 \\
    PSNR &\textbf{32.8} &28.7 &12.0 \\
    \bottomrule
  \end{tabular}
\end{wraptable}
  
\subsubsection{Role of finetuning}
\label{subsubsec:ablation_finetune}
\begin{figure}[t]
  \centering
  \includegraphics[width=\linewidth]{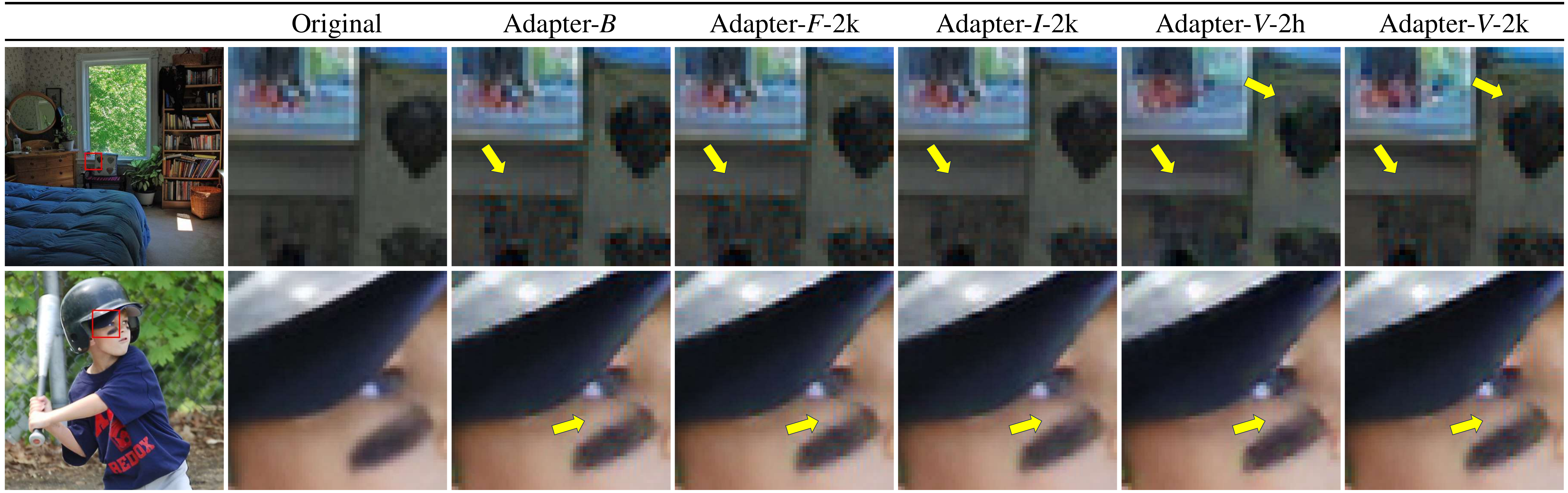}
   \caption{Qualitative comparison between different finetuning strategies on COCO~\citep{lin2014microsoft} images. Adapter-{\it B} and \oursf~produces tiny grid-like artifacts. Finetuning with VAE (\oursi~and -{\it V}) alleviates this issues. Using fintuned VAE at inference time (Adapter-{\it V}) leads to blurry images. Using original VAE (\oursi) achieves the most visually appealing results. Zoom in for best view.}
   \label{fig:ablation_finetuning}
   \vspace{-0.2cm}
\end{figure}

Tab.~\ref{tab:ablation_finetuning} and Fig.~\ref{fig:ablation_finetuning} compare different finetuning strategies quantitatively and qualitatively.
From Tab.~\ref{tab:ablation_finetuning}, we can see that Adapter-{\it B} achieves good numerical results. However, when we zoom in on the generated images by 20x times, tiny grid-like artifacts become visible. If we freeze the VAE decoder and perform a quick fine-tuning for 2k steps using a large learning rate, resulting in Adapter-{\it F}-2k. We find that while the accuracy remains almost unchanged, the PSNR and SSIM metrics improve, but the artifacts are still there.

If we employ the Hybrid Finetuning (\oursi), we can find that the artifacts are effectively removed. Additionally, because the VAE remains intact, this results in the sharpest and most visually appealing images, with the PSNR significantly improving to 34.8 dB. However, this comes at a little cost of a 0.02 loss in bit accuracy under combined attacks.

If we adopt the Joint Finetuning (Adapter-{\it V}), we observe an additional 0.01 improvement in bit accuracy compared to the baseline. However, this approach results in a decline in PSNR, SSIM, and FID. Visual comparisons in Fig.~\ref{fig:ablation_finetuning} show that Adapter-{\it V} produces smoother and blurrier images, which accounts for the decrease in metrics and also the elimination of the tiny grid-like artifacts seen with Adapter-{\it B}. As the fine-tuning steps increase from 200 to 2k, the images become sharper, but the grid-like artifacts re-emerge.
Notably, using the fine-tuned VAE during inference introduces lens flare-like artifacts, as shown in Fig.~\ref{fig:cmp_stablesig} and discussed in Sec.~\ref{subsec:qualitative}. 

In summary, finetuning can further enhance image quality or accuracy, but different finetuning strategies introduce subtle trade-offs between accuracy, image sharpness, and artifacts. Considering the plug-and-play nature of adapters, we adopt Adapter-{\it F}-2k and Adapter-{\it I}-2k as our default choices, as they do not require a fine-tuned VAE decoder for inference.


\begin{table}
  \caption{Comparison between different finetuning strategies. "Adapter-{\it B}" means no extra finetuning. "-2k" and "-2h" means finetuning with extra 2000 and 200 steps, respectively. Bit Acc is evaluated under combined attacks.}
  \label{tab:ablation_finetuning}
  \centering
  \begin{tabular}{lllll}
    \toprule
     & Bit Acc &PSNR &SSIM &FID \\
    \midrule
    Adapter-{\it B} & 0.93 & 32.8 & 0.94 & 2.7 \\
    \midrule
    Adapter-{\it F}-2k & 0.93 & 33.0 & 0.95 & 2.7 \\
    Adapter-{\it I}-2k & 0.91 & \textbf{34.8} & \textbf{0.96} & \textbf{2.5} \\
    Adapter-{\it V}-2h & \textbf{0.94} & 29.1 & 0.86 & 3.9 \\
    Adapter-{\it V}-2k & \textbf{0.94} & 29.9 & 0.87 & 3.1 \\
    \bottomrule
  \end{tabular}
  \vspace{-0.3cm}
\end{table}

\subsection{Qualitative Results}
\label{subsec:qualitative}
\begin{figure}[ht]
  \centering
  \includegraphics[width=\linewidth]{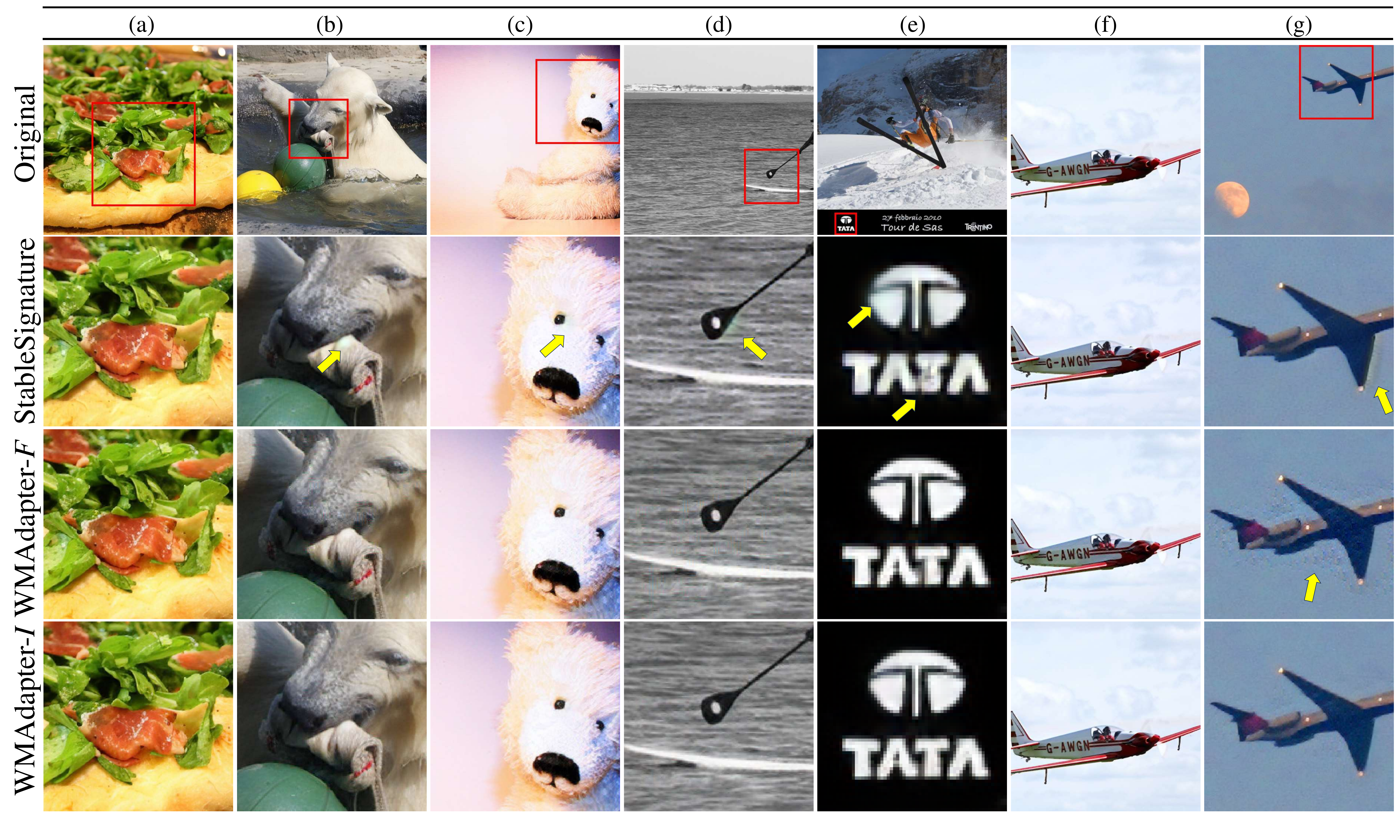}
   \caption{Comparison between WMAdapter and StableSignature~\citep{fernandez2023stable}. Yellow arrows point to the generated artifacts. Best viewed in color and zoomed in.}
   \label{fig:cmp_stablesig}
\end{figure}

We qualitatively compare our WMAdapter with Stable Signature~\citep{fernandez2023stable} in Fig.~\ref{fig:cmp_stablesig}. We can find that both Stable Signature and ours perform well on the image with rich textures (column {\it a}). However, we observe that Stable Signature tends to produce green/yellow spot-like artifacts (columns {\it bcdeg}) as pointed by the yellow arrows. This may be a side effect by finetuning the VAE decoder. As we did not observe such artifacts in \oursf, but found slight signs in Adapter-I, which also finetunes the VAE decoder. Since \oursi~uses the original VAE decoder during inference, artifacts are greatly suppressed. This spot-like artifact can also be observed in other works~\citep{xiong2023flexible} that also finetune the VAE decoder. By comparing column (\textit{e}), we can see that our adapters produce sharper images, which is also supported by the PSNR metric. By comparing columns (\textit{df}), we can see that Stable Signature has a slight reddish tint. Our adapters keep the tint more accurately. Please also see the "Green Apple" example in Fig.~\ref{fig:appendix_difference_others}. Column (\textit{g}) shows the failure case of \oursf. It creates artifacts around the plane. This may happen when the background and foreground objects are of uniform color with little textures. In short, compared with StbaleSignature that finetunes VAE decoder to embed watermarks, WMAdapter can generate sharper images with higher watermark invisibility.

\vspace{-0.2cm}

\section{Conclusion and limitation}
\label{sec:conclusion}

In this paper, we present the WMAdapter, a plug-and-play watermark plugin that enables latent diffusion models to embed arbitrary bit information during image generation. Our adapter is lightweight, easy to train, and offers greater flexibility, comparable robustness, and higher image quality compared to previous diffusion watermarking counterparts due to its dedicated design. However, we find one variant \oursf ~tends to produce noticeable artifacts when watermarking images with homogenous background colors.  In summary, WMAdapter provides a simple yet effective baseline for further exploration on diffusion watermarking.


\bibliographystyle{ieeenat_fullname}
\bibliography{main}

\clearpage

\clearpage

\appendix

\section{Appendix}

\subsection{Experiment statistical significance}
\label{sec:appendix_statistical_significance}
For the first training stage, we ran 3 independent training and found the standard deviation of average validation bit accuracy across 3 runs to be $0.0006$, and the standard deviation of validation PSNR to be $0.03$ dB. 

For the second finetuning stage, we also ran 3 independent trials. The standard deviation of average validation bit accuracy across 3 runs was also $0.0006$, and the std of validation PSNR was $0.04$ db. The small standard deviation at both stages demonstrates the stability of our method. Since the standard deviation is too small to be clearly viewed in Fig.~\ref{fig:appendix_range}, we report the numbers in text.

\subsection{Broader impacts}
\label{sec:appendix_broader_impacts}
The proposed diffusion watermarking technique offers significant positive societal impacts, such as enhancing copyright protection for digital creators and helping to prevent the spread of fake news by enabling the authentication of images. However, it also poses potential negative impacts, including privacy concerns, the risk of misuse for malicious purposes, technical challenges that may disadvantage smaller creators, and possible degradation of image quality. Balancing these benefits and drawbacks is crucial to ensure the responsible and effective use of this technology.

In terms of applications, our proposed WMAdapter can also be directly applied to video generation models such as AnimateDiff~\citep{guo2023animatediff} and StableVideoDiffusion~\citep{blattmann2023stable}, which share the same VAE architecture as image Diffusion models. We leave further exploration on video to the future work.

\subsection{Evaluation on various distortion intensities}
\begin{figure}[t]
  \centering
  \includegraphics[width=\linewidth]{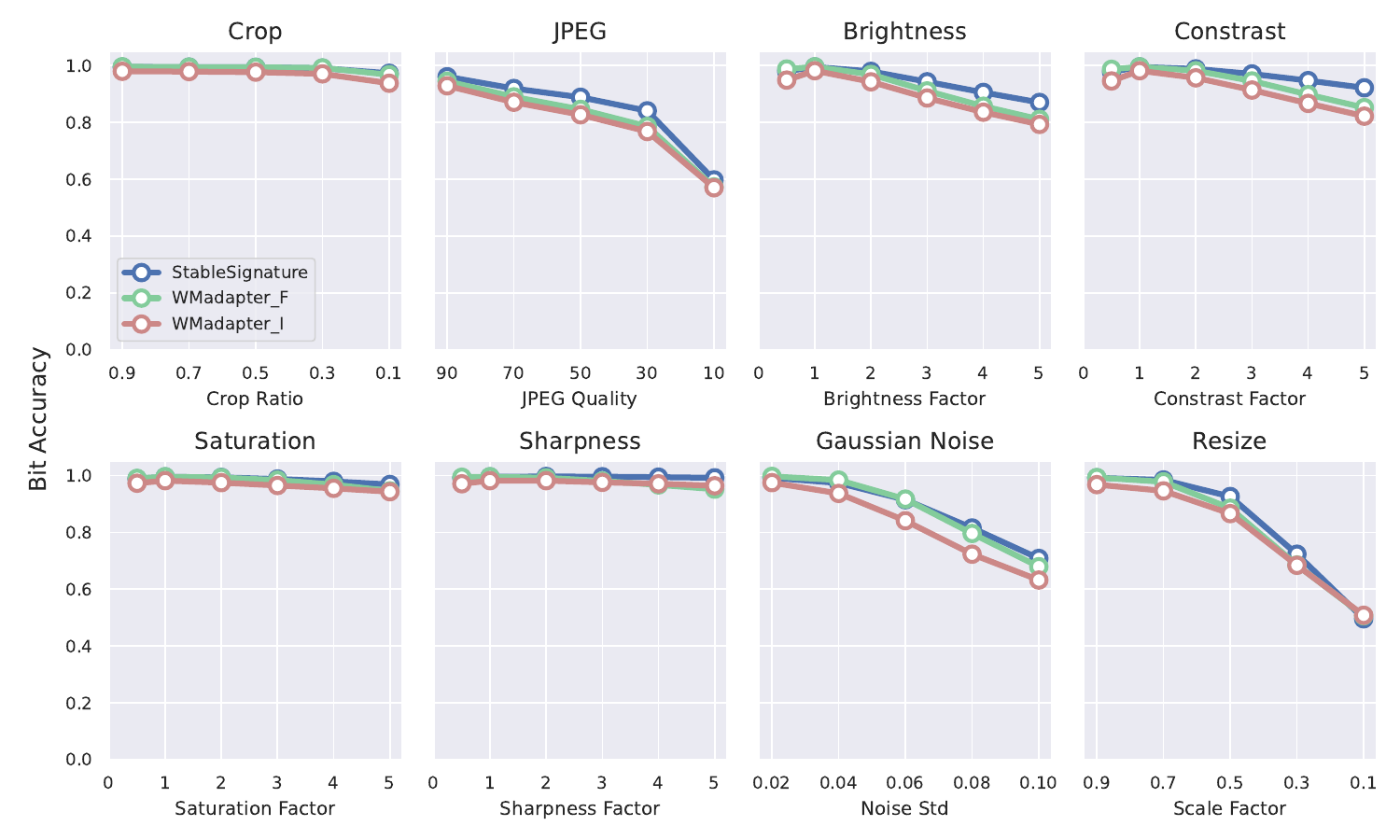}
   \caption{Robustness comparison over various distortion intensities.}
   \label{fig:appendix_range}
\end{figure}

Fig.~\ref{fig:appendix_range} evaluates our method under larger ranges of distortion intensities and more attacks. We can see that our adapters remain comparable robustness to Stable Signature~\citep{fernandez2023stable} over range of attack intensities. Note that all three methods exhibit limited robustness to significant Gaussian noise. This limitation arises because the pretrained HiDDeN checkpoint~\citep{fernandez2023stable} was not specifically trained to handle noise attacks. To provide a comprehensive evaluation of the different methods, we still present their results under Gaussian noise.

\subsection{Results on different VAEs}

We train several watermark adapters for VAEs used by SD1.5\&2.1~\citep{Rombach_2022_CVPR}, SDXL~\citep{podell2023sdxl} and DiT~\citep{peebles2023scalable} (kl-f8-mse) at resolution $512 \times 512$. We compare the adapters before the finetuning stage. Tab.~\ref{tab:appendix_other_vaes} shows the results. We observe that WMAdapter consistently performs well across various VAEs, making it applicable in a wide range of contexts. The PSNR of SDXL adapter is lower compared to SD2.1 and DiT VAE. This may be caused by the resolution mismatch.
\begin{table}[ht]
  \caption{Evaluation on VAEs from different models.}
  \label{tab:appendix_other_vaes}
  \centering
  \begin{tabular}{lllll}
    \toprule
     &SD1.5 &SD2.1 &SDXL &DiT \\
    \midrule
    Bit Acc &0.99 &0.99 & 0.99 & 0.99 \\
    PSNR & 32.1 & 32.8 & 31.2 & 32.4 \\
    \bottomrule
  \end{tabular}
\end{table}

We further evaluate the ability of WMAdapter to zero-shot transfer to different VAEs. Specifically, we apply the adapter trained on SD2.1 directly to SD1.5 VAE and find that it is able to handle SD1.5 image latents with little performance loss. This empirical result demonstrates the zero-shot transfer potential of WMAdapter to different customized SD VAEs.

\subsection{Visualization of attacks}
\begin{figure}[t]
  \centering
  \includegraphics[width=\linewidth]{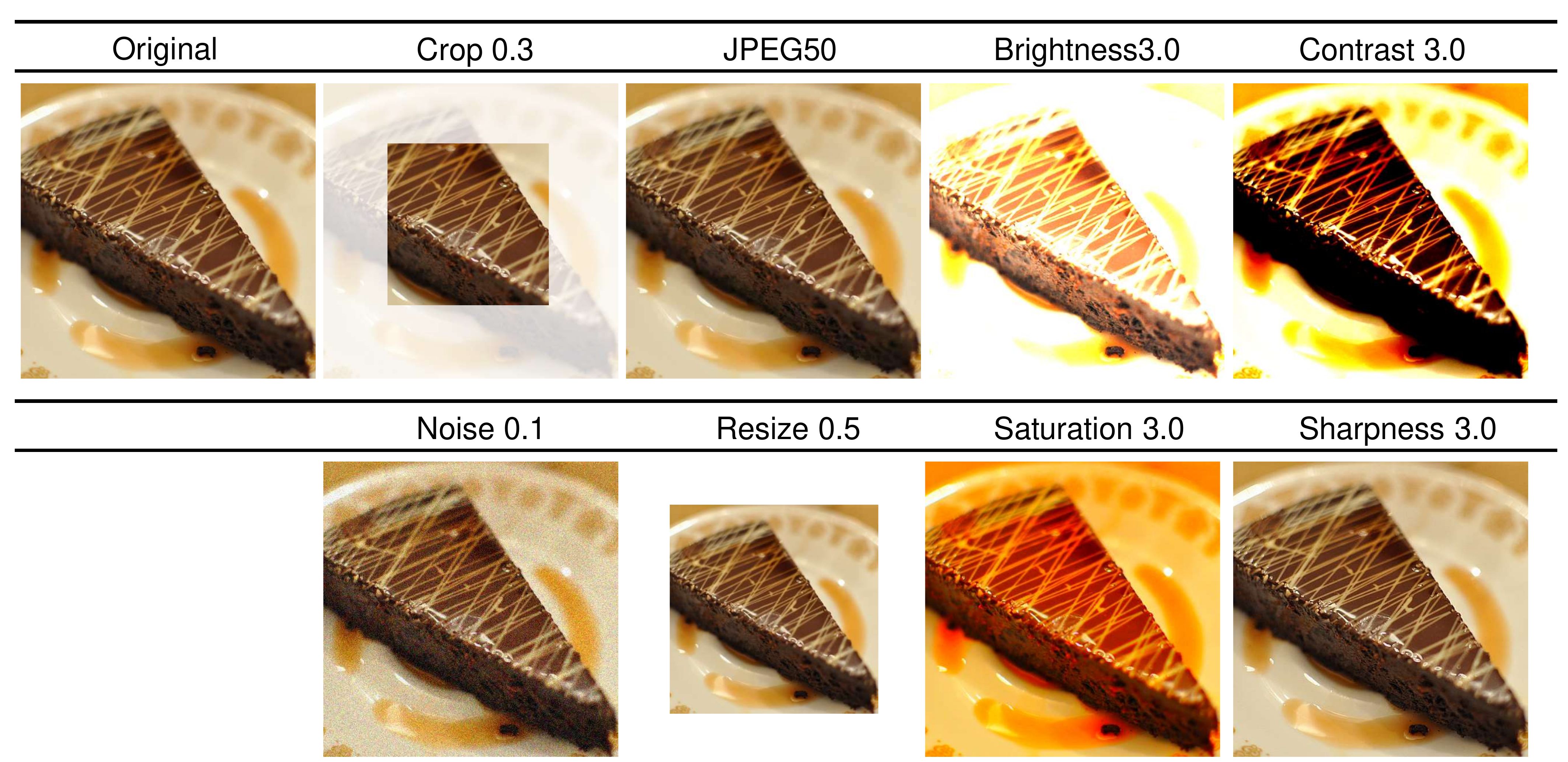}
   \caption{Visualization of different augmentations.}
   \label{fig:appendix_augmentation}
\end{figure}

Fig.~\ref{fig:appendix_augmentation} shows different attacks evaluated in the paper. 

\subsection{More qualitative results on COCO dataset}
\begin{figure}[ht]
  \centering
  \includegraphics[width=\linewidth]{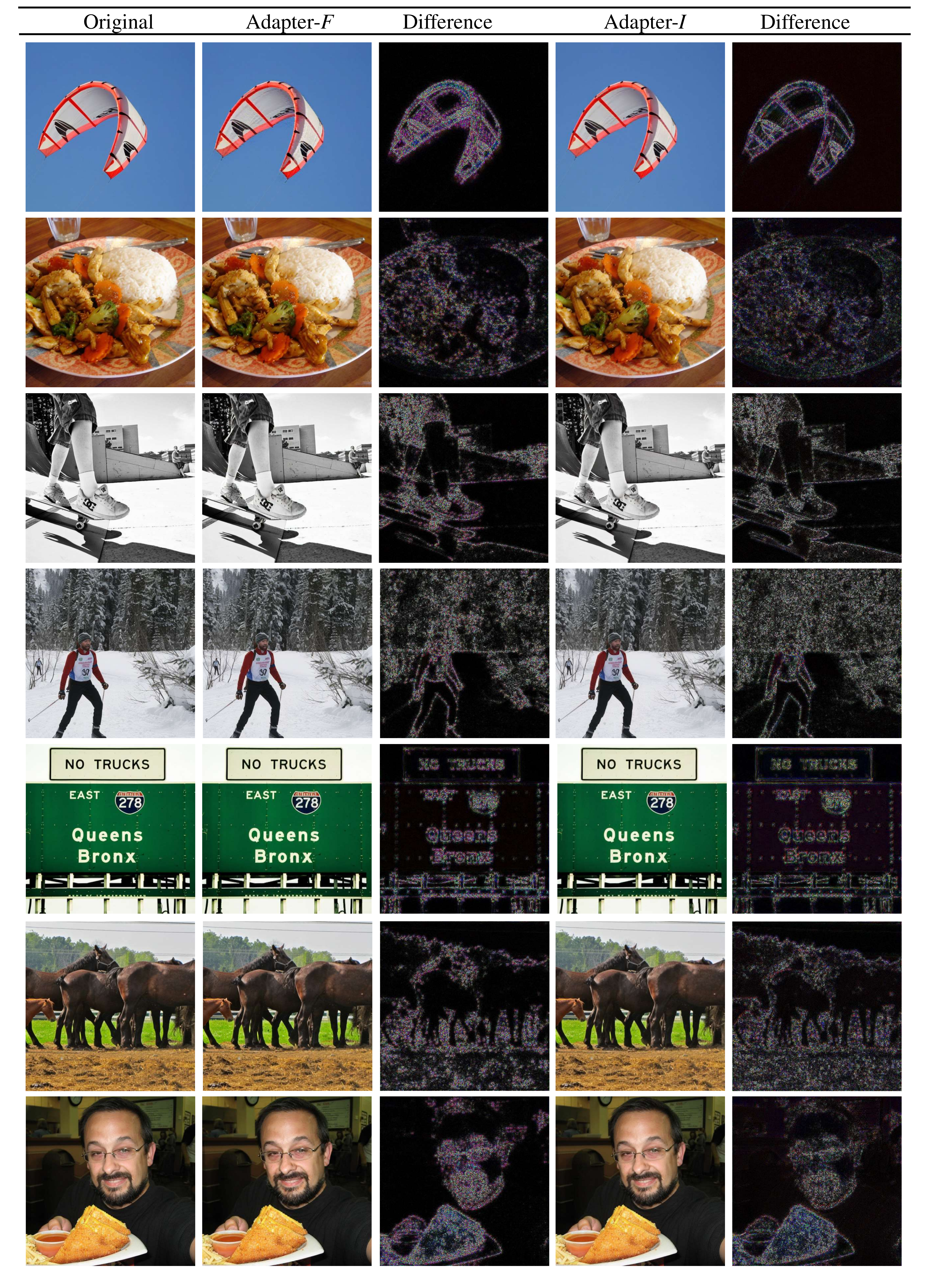}
   \caption{Qualitative results on COCO dataset at resolution 512.}
   \label{fig:appendix_difference}
\end{figure}

Fig.~\ref{fig:appendix_difference} shows the watermarked images and their difference with the original images. We find that both \oursff~and \oursii~can adaptively embed watermark information into regions with significant color variations and richer textures in the images, significantly enhancing their invisibility.

\subsection{Generalization to Ideogram dataset}
\begin{figure}[ht]
  \centering
  \includegraphics[width=\linewidth]{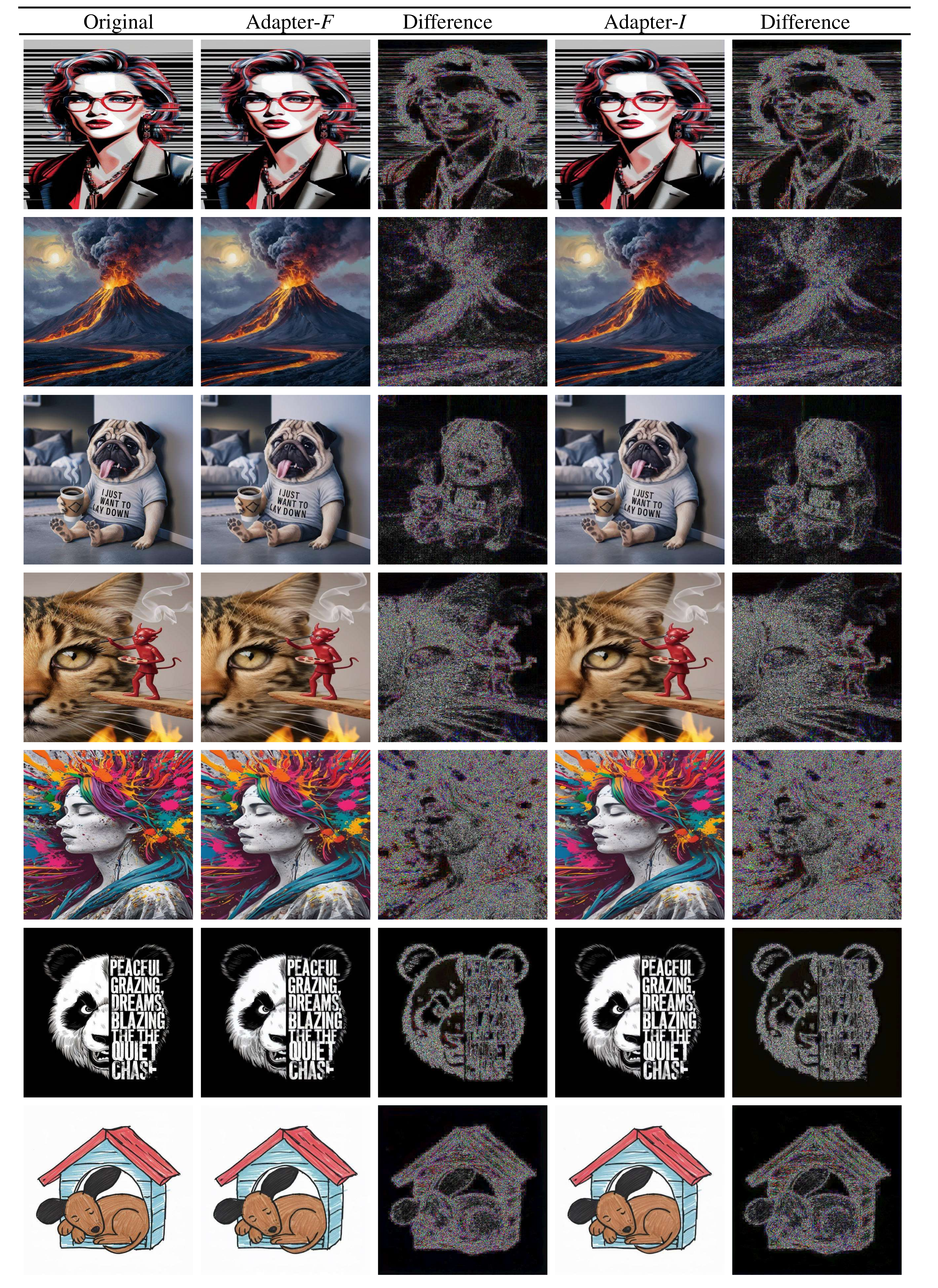}
   \caption{Qualitative results on Ideogram~\citep{ideogram} at resolution 512.}
   \label{fig:appendix_difference_ideogram}
\end{figure}

Fig.~\ref{fig:appendix_difference_ideogram} shows our results on images generated by Ideogram~\citep{ideogram}. These images exhibit completely different styles. However, our WMAdapter, trained on COCO, transfers seamlessly to them.

\subsection{Comparison with other methods}
\begin{figure}[ht]
  \centering
  \includegraphics[width=\linewidth]{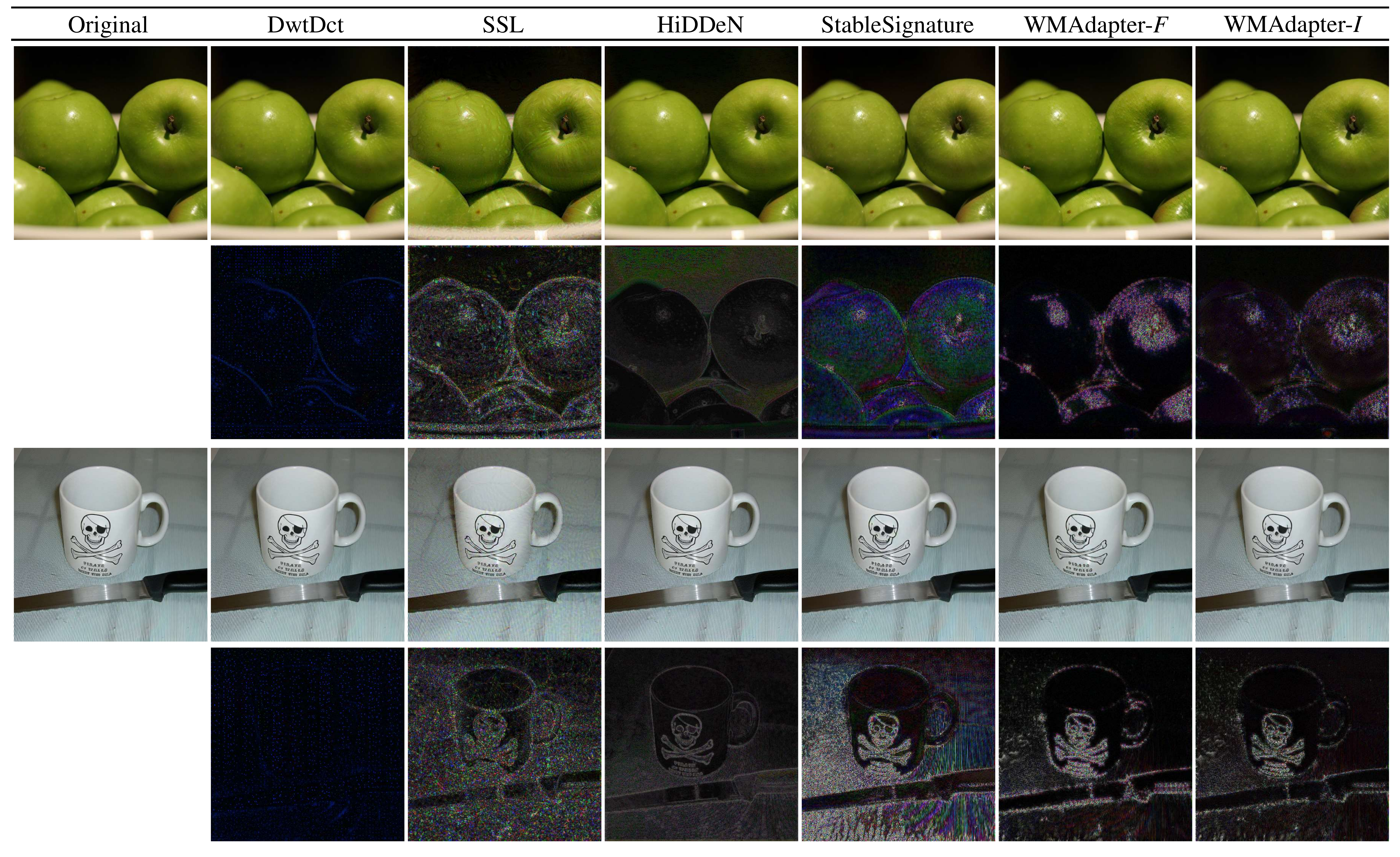}
   \caption{Qualitative comparison on MS-COCO with other invisible watermarking methods. For HiDDeN results~\citep{zhu2018hidden}, we use the model trained by~\citep{fernandez2023stable} with inference time modulation by a perceptual just noticeable difference (JND) mask to enhance invisibility~\citep{fernandez2022active}. Zoom in for best view.}
   \label{fig:appendix_difference_others}
\end{figure}

Fig.~\ref{fig:appendix_difference_others} compares different watermarking methods on COCO~\citep{lin2014microsoft} val images. We observe that the traditional method DwtDct~\citep{cox2007digital} introduces minimal modifications to the original image, which consequently makes it sensitive to various attacks. SSL~\citep{fernandez2022watermarking} produces visible artifacts that significantly degrade visual quality. Following~\citep{fernandez2023stable}, we employ the post-hoc JND (Just Noticeable Difference) mask~\citep{fernandez2022active} for HiDDeN~\citep{zhu2018hidden} to enhance its invisibility. Note that HiDDeN imprints watermark perturbation even to uniform black backgrounds in images ‘Apple’. In contrast, our approach automatically learns to apply stronger modifications only to edges or texture-rich regions, achieving results similar to the post-hoc mask. Importantly, neither of our adapters attempts significant alterations in the black background of the ‘Apple’ image.

\begin{figure}[ht]
  \centering
  \includegraphics[width=\linewidth]{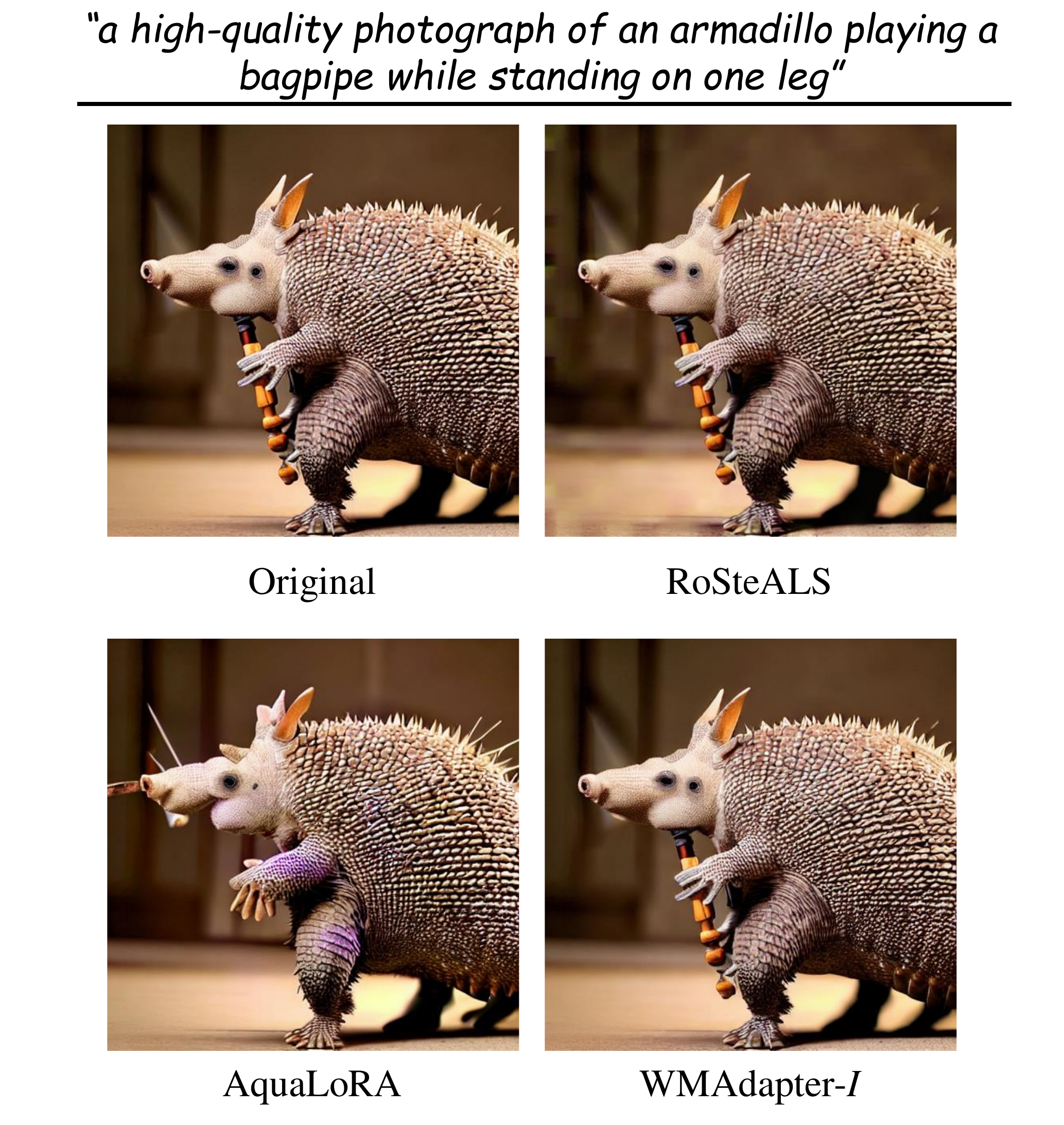}
   \caption{Qualitative comparison with other diffusion watermark plugins.}
   \label{fig:appendix_other_plugins_1}
\end{figure}

\begin{figure}[ht]
  \centering
  \includegraphics[width=\linewidth]{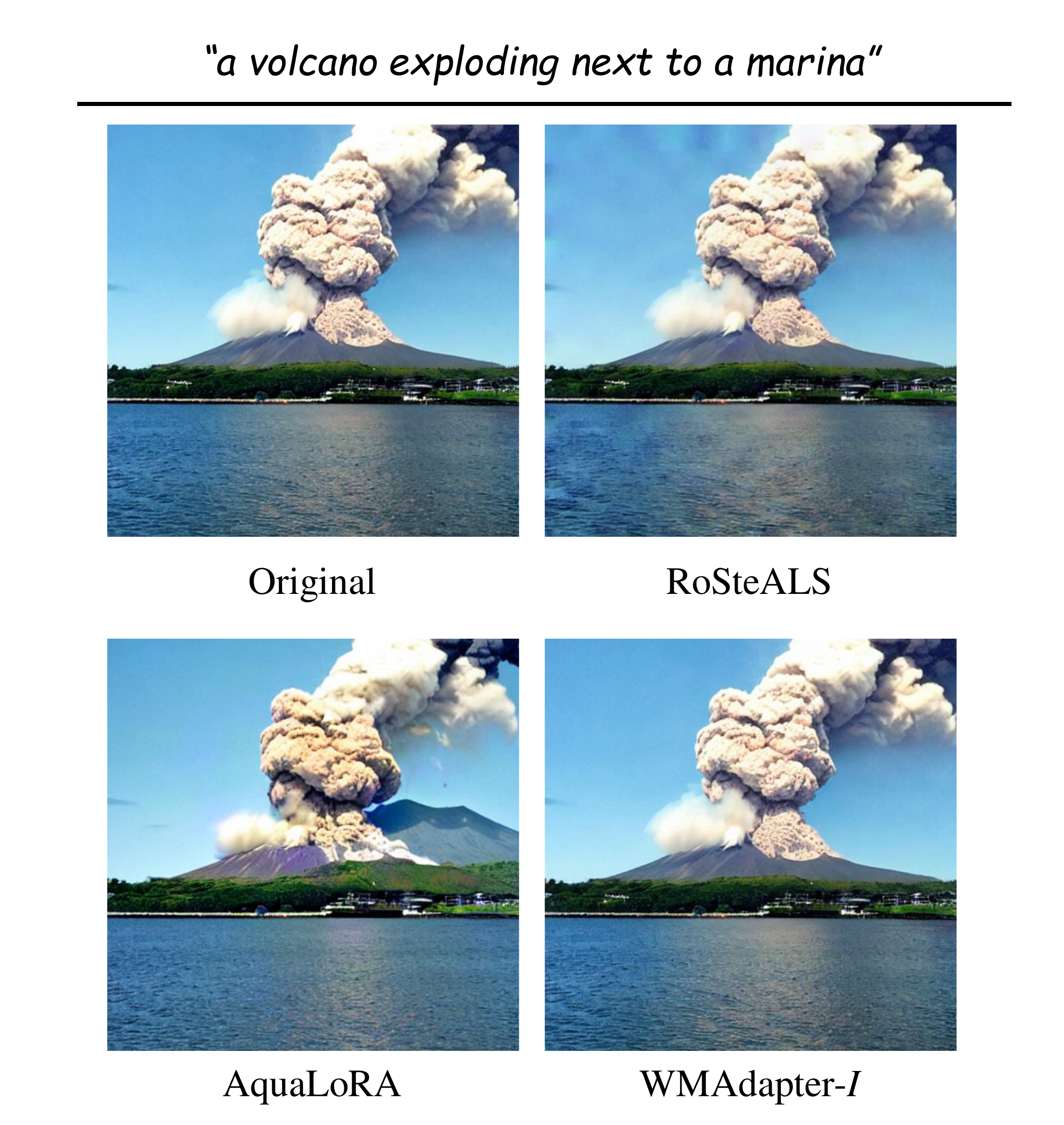}
   \caption{Qualitative comparison with other diffusion watermark plugins.}
   \label{fig:appendix_other_plugins_2}
\end{figure}

Fig.~\ref{fig:appendix_other_plugins_1} and Fig.~\ref{fig:appendix_other_plugins_2} compares WMAdapter and two recent watermark plugin methods RoSteALS~\citep{bui2023rosteals} and AquaLoRA~\citep{feng2024aqualora} on SD generated images. We can see that  RoSteALS tends to produce visible patch-like artifacts and AquaLoRA introduces visible purple biases. WMAdapter produces images that are most consistent with the original images and have the least artifacts.

\end{document}